\documentclass[journal]{IEEEtran}

\usepackage[utf8]{inputenc} 
\usepackage[T1]{fontenc}    
\usepackage[hidelinks]{hyperref}       
\usepackage{url}            
\usepackage{booktabs}       
\usepackage{amsfonts}       
\usepackage{graphicx}
\usepackage{amsmath}
\usepackage{algorithm}
\usepackage{algorithmic}
\usepackage{xcolor}
\usepackage{flushend}
\usepackage{threeparttable}

\makeatletter
\newcommand{\removelatexerror}{\let\@latex@error\@gobble}
\makeatother

\newcommand{\convruntime}{9.4}
\newcommand{\convto}{97.9}
\newcommand{\convtt}{99.7}
\newcommand{\rnnruntime}{48.1}
\newcommand{\rnnto}{96.2}
\newcommand{\rnntt}{99.6}
\newcommand{\tablesize}{\scriptsize}
\newcommand{\tableconv}{Table A.\ref{table:conv}}
\newcommand{\tablernn}{Table A.\ref{table:rnn}}

\begin{document}
\title{Accelerating Neural ODEs Using Model Order Reduction}


\author{Mikko Lehtimäki, Lassi Paunonen, Marja-Leena Linne
\thanks{Mikko Lehtimäki and Marja-Leena Linne are with the
    Faculty of Medicine and Health Technology,
    Tampere University, Finland.
    Lassi Paunonen is with the Faculty of Information Technology and
    Communication Sciences, Tampere University, Finland. }%
\thanks{Correspondence: mikko.lehtimaki@tuni.fi}}


\hypersetup{
pdftitle={DEIMNet paper},
pdfsubject={DEIMNet},
pdfauthor={Mikko Lehtimäki},
pdfkeywords={Neural ODEs, pruning, compression, acceleration, Proper Orthogonal
Decomposition, Discrete Empirical Interpolation Method},
}

\markboth{Preprint. This work has been submitted to the IEEE for possible
publication. Copyright may be transferred without notice.}%
{Shell \MakeLowercase{\textit{et al.}}: IEEEtran.cls for IEEE Journals}

\maketitle

\begin{abstract}

Embedding nonlinear dynamical systems into artificial neural networks is a
powerful new formalism for machine learning. By parameterizing ordinary
differential equations (ODEs) as neural network layers, these Neural ODEs are
memory-efficient to train, process time-series naturally and incorporate
knowledge of physical systems into deep learning models. However, the practical
applications of Neural ODEs are limited due to long inference times, because
the outputs of the embedded ODE layers are computed numerically with
differential equation solvers that can be computationally demanding.  Here we
show that mathematical model order reduction methods can be used for
compressing and accelerating Neural ODEs by accurately simulating the
continuous nonlinear dynamics in low-dimensional subspaces. We implement our
novel compression method by developing Neural ODEs that integrate the necessary
subspace-projection and interpolation operations as layers of the neural
network. We validate our approach by comparing it to neuron pruning and
SVD-based weight truncation methods from the literature in image and
time-series classification tasks. The methods are evaluated by
acceleration versus accuracy when adjusting the level of compression. On this
spectrum, we achieve a favourable balance over existing methods by using model
order reduction when compressing a convolutional Neural ODE. In compressing a
recurrent Neural ODE, SVD-based weight truncation yields good performance.
Based on our results, our integration of model order reduction with Neural ODEs
can facilitate efficient, dynamical system-driven deep learning in
resource-constrained applications.

\end{abstract}

\IEEEpeerreviewmaketitle

\section{Introduction}

Deep Learning (DL) is reaching and surpassing human performance in
domain-specific applications~\cite{lecun2015deep}. Accordingly, there is
increased demand for including DL-based algorithms into consumer devices that
may contain only limited computational capacity and be battery-powered, making
resource efficiency of DL-based algorithms important.
An interesting new development in DL research is Artificial Neural Networks
(ANNs) that employ dynamical systems, replacing traditional discrete layers
with a continuous-time layer in the form of ordinary differential equations
(ODEs)~\cite{weinan2017proposal, lu2018beyond, haber2017stable,
chen2018neural}. In these Neural ODEs, the continuous formalism allows flexible
processing of time-series and irregularly sampled data, such as medical and
physiological signals~\cite{rubanova2019latent}.  Moreover, Neural ODEs are
useful for resource-constrained and embedded applications because they are very
memory and parameter efficient~\cite{chen2018neural}. The ODE layer also
facilitates engineering physical details, such as energy conservation laws or
spectral properties, into neural networks~\cite{haber2017stable}. However,
often a big computational bottleneck in Neural ODEs is the dynamical system
layer, since propagating data through the system requires many evaluations
using numerical ODE solvers.  Reducing the computational cost of the ODE layer
is the main motivation of our work.

In this work we show that Neural ODEs can be accelerated by compressing them
using model order reduction (MOR) methods. The MOR approach is based on
projecting dynamical systems onto low-dimensional subspaces. Here, we develop
MOR methods that are integrated into Neural ODEs. In this manner, we lower the
required storage size, memory consumption, multiply-adds and nonlinear
activation count needed to compute predictions from input data. The resulting
compressed Neural ODEs can be deployed for real-time computing and devices
where energy efficiency is paramount. In order to validate the performance of
our MOR method, we compare it to two established compression methods from the
literature. Our results demonstrate that MOR is 
a theoretically grounded and effective method for accelerating
Neural ODEs, because it achieves a favourable, adjustable and extensible
balance between speedup and accuracy of compressed models. We show this result
in two different classification tasks that use different Neural ODE
architectures, a convolutional and a recurrent neural network (RNN).


Compressing ANNs is one of the principal ways of converting high-performing
trained networks into more efficient networks, since good accuracy can often be
recovered without long training times, while the size of the compressed network
can be chosen in a flexible manner. Several neural network compression methods
have been proposed for accelerating ANNs~\cite{blalock2020state,
han2015learning, hu2016network}.  These include singular value decomposition
(SVD) based weight truncation~\cite{denton2014exploiting, girshick2015fast} and
neuron pruning with an importance score based on zero
activations~\cite{hu2016network}, which we implement for comparison to our MOR
method. Many prominent existing approaches rely on assigning importance scores
to neurons based on their connection weights or output values. However, we
argue that such methods are non-optimal for compressing Neural ODEs, as it is
difficult to quantify the importance of neurons in the ODE layer based on
criteria such as the output of the layer alone. This is because the final state
of the nonlinear ODE system gives no information about the dynamics of the
actual computations. In contrast, MOR methods are designed precisely for the
approximation of state trajectories and input-output relationships in dynamical
systems, hence they can overcome the shortcomings of existing compression
methods when aiming to accelerate Neural ODEs. 

Our approach to model acceleration is inspired by computational neuroscience.
Due to high computational burden, modeling studies of the brain in realistic
scales are limited to simulating small fractions of the brain using
supercomputers~\cite{Markram2015}. To overcome computational resource
challenges, MOR methods have been adapted successfully for reducing single
neuron~\cite{Kellems2010, Du2014}, synapse~\cite{lehtimaki2017order} and
neuronal population~\cite{lehtimaki2019projection, lehtimaki2020accelerated}
models. Moreover, the similarities in models of the brain as well as artificial
neural networks, which include connectivity patterns and nonlinear computation
units, make us hypothesize that MOR methods may be a principled approach for
accelerating Neural ODEs. 

Our MOR approach for compressing Neural ODEs is based on the Proper Orthogonal
Decomposition (POD)~\cite{Lumley1993} with the Discrete Empirical Interpolation
Method (DEIM)~\cite{Chaturantabut2010}, a variant of the Empirical
Interpolation Method~\cite{barrault2004empirical}. Using the POD-DEIM method,
we derive reduced order models (ROMs) that can be simulated efficiently in
low-dimensional subspaces, even in the presence of nonlinear activation
functions. In the context of Neural ODEs, our method compresses the ODE
block by projecting it onto a low-dimensional subspace using POD, which
reduces the number of state variables in the ODE system. Linear operations
are compressed by transformation into this subspace. When nonlinear activation
functions are present, using DEIM we determine the most informative output
neurons based on their time dynamics, and prune the other neurons. This reduces
the dimensionality of the nonlinear operation and removes rows from the weight
matrix, since connection weights of pruned neurons are discarded. We
interpolate an approximate response for the pruned neurons directly in the
low-dimensional subspace. In convolutional layers, the reduced model computes
convolutions only at the selected interpolation points so that every kernel has
its own set of evaluation coordinates.

We integrate the subspace projection and interpolation steps of POD-DEIM into
the Neural ODEs as layers, and this introduces additional operations into the
neural network (see Figure~\ref{fig:deimnet}). In some network architectures
these steps actually further reduce the overall number of multiply-adds in the
neural network. The POD-DEIM reduction is applied after training the model and
the reduced model can be fine-tuned for increased accuracy. In summary, Neural
ODEs allow us to bridge a gap between ANN research and control theory research,
so that a substantial amount of previously unemployed knowledge in model
reduction can be utilized in ANN compression and acceleration. For example,
analytical optimality results and error bounds exist for our MOR
algorithms~\cite{Lumley1993, Chaturantabut2012}. 

In Section~\ref{sec:related} we review previous work in compressing and
accelerating neural networks in general and Neural ODEs specifically.  In
Section~\ref{sec:methods} we present our proposed model order reduction
approach and show how to formulate it in the context of continuous neural
networks. In the same section we introduce two established acceleration methods
that we compare our method to in benchmark problems. In
Section~\ref{sec:results} we provide theoretical compression ratios for the
chosen methods and show actual accuracy and wall-time metrics of compressed
Neural ODEs in two different classification tasks, one using a convolutional
and the other using a recurrent ODE architecture. We discuss the success of the
model reduction approach and the significance of our work in
Section~\ref{sec:discussion} with future suggestions and conclude in
Section~\ref{sec:conclusion}.

\section{Related work}\label{sec:related}

Several studies have reported that ANNs contain structures and redundancies
that can be exploited for reducing memory and power
usage~\cite{blalock2020state}.  Here we focus on structural acceleration
approaches that modify trained networks to achieve a more efficient
architecture, and leave efforts such as low precision algebra, quantization of
weights~\cite{vanhoucke2011improving},
binarization~\cite{courbariaux2015binaryconnect},
hashing~\cite{chen2015compressing}, vectorization~\cite{ren2015vectorization},
frequency space evaluation~\cite{mathieu2014fast} and
adjusting ODE solver tolerances or step size~\cite{chen2018neural} out of the
present study since those are complementary to the approaches presented here.
Moreover, several hardware accelerators have been proposed (for
example~\cite{chen2014diannao}), and those will not be addressed here.
Furthermore, computational bottlenecks in ANNs have also been addressed by
first reducing data dimension and then training simpler models.  These
techniques range from feature engineering to data dimensionality reduction.
However, in this work our focus is on compressing trained \textit{networks},
and hence \textit{data} compression is considered complementary to out
approach. Structural compression methods can be further grouped into several
categories.  

\paragraph{Pruning weights} Prior work has addressed weight
pruning~\cite{lecun1990optimal, hassibi1993second} and enforcing weight
sparsity~\cite{yu2012exploiting, xue2013restructuring} during training to
obtain weight matrices that require less storage space and memory than dense
weight matrices. Several methods to evaluate weight importance have been
proposed, see for example a recent review addressing 81 pruning
studies~\cite{blalock2020state} and compares the achieved compression rates and
speedups for several ANN models. It is common to include pruning in the
training loop, because altering weights after training leads to accuracy loss.
In order to achieve significant acceleration with weight-pruning methods the
use of special software, masking strategies, sparse algebra or hardware
accelerators is recommended~\cite{han2016eie, parashar2017scnn}.

\paragraph{Decomposition} Decompositions and low-rank matrix factorizations
have been used for compress network weights so that linear operations in a
layer are computed efficiently~\cite{rigamonti2013learning,
denton2014exploiting, jaderberg2014speeding, gong2014compressing,
girshick2015fast}. Decomposition is based on the
observation that weight matrices, especially as their size increases, are
seldom full rank. The idea is that a fully connected weight matrix or a
tensorized convolutional kernel can be decomposed into compact low-rank
matrices that require less storage space, memory and multiply-adds during
training and testing. An approximation of the original operation is then
obtained as a product of the low-rank tensors. In addition to compressing
linear operations, prior work has also addressed decomposition by taking the
nonlinear response into account~\cite{zhang2015accelerating}.  The cost of
decomposition approaches is that a single layer is replaced with multiple
smaller ones, which trades parallelism for forward operations, canceling out
some of the obtained acceleration. 

\paragraph{Pruning neurons} Weight pruning and decomposition approaches
maintain the structure of the compressed network in that the number of
nonlinear activation functions is not changed. Eliminating activation
functions leads to acceleration as an entire row of weights from a fully
connected layer can be removed~\cite{han2015learning, van2015cross}. As neuron
pruning changes the number of neurons in a layer, the next layer must take this
into account, for example by deleting columns (inputs) from the weight matrix
in the case of fully connected layers. Alternatively, interpolation can be used
to approximate the original response, possibly introducing additional
computation. In fully connected architectures large compression rates in
storage space and memory are obtained when rows or columns are deleted from
fully connected layers. Importance scores, such as percent of zero activations
have been developed for identifying prunable neurons~\cite{hu2016network}.
Pruning and quantization have been combined with Huffman coding into a
compression framework that delivers memory and energy
efficiency~\cite{han2016deep}. An optimization approach has been used to
enforce sparse columns in a fully connected layer so that the corresponding
input neurons can be pruned~\cite{van2015cross}. 

\paragraph{Pruning filters} Convolutional Neural Networks (CNNs) have attracted
a lot of attention in the compression literature. This is not surprising given
their good rate of success in real-life tasks and their high computational
cost.  Convolutional operations make the bulk of modern image and video
processing networks and are used in many other applications, such as sequence
processing.  It is possible to approach CNN compression by analysing either the
convolutional filters or feature maps obtained by applying the filters on input
data~\cite{he2017channel, molchanov2017pruning, lebedev2018speeding}. The
filters can be compressed with decomposition methods, similarly to fully
connected layers.  Alternatively, entire kernels can be pruned from filters.
Pruning kernels is very effective as intermediate feature maps are eliminated.
The number of methods and criteria available for identifying pruning targets in
CNNs is very high~\cite{lebedev2018speeding}. In convolutional networks, neuron
pruning corresponds to skipping a kernel evaluation at a single spatial
location.  However, such an operation is rarely supported by deep learning
frameworks, and pruning in convolutional networks has focused on eliminating
parts of or entire filters or feature maps. 

In the context of Neural ODEs, a few studies have addressed the
acceleration of inference and training times. One approach focuses on
learning simple dynamics that do not burden ODE solvers as much as stiff
systems~\cite{dupont2019augmented, kelly2020learning, finlay2020train}.
Additionally, training times have been reduced by further improving the
adjoint method, for example by relaxing error criteria~\cite{kidger2020hey}
while inference times have been accelerated by introducing hypersolvers -
neural networks that solve Neural ODEs~\cite{poli2020hypersolvers}. These
methods are complementary to ours, since we aim at compressing the learned
architecture with MOR methods.  Our approach is comparable to pruning
neurons based on an importance score~\cite{han2015learning, hu2016network},
so that a number of rows from the weight matrix and nonlinear activation
functions can be removed altogether. The POD-DEIM method improves on the
existing methods of determining neuron importance, as it accounts for the
complete dynamics of the ODE block. Additionally, POD-DEIM can be combined
with many of the prior methods such as quantization of weights.

\section{Methods}\label{sec:methods}

In this section, we present the theory of the continuous interpretation of
ANNs.  We then describe our model order reduction method and show how to
formulate it in the ANN and Neural ODE context. Finally, we present two other
ANN acceleration methods from the literature, which we use as benchmarks to our
method, and their implementation in the Neural ODE setting.

\subsection{Continuous networks}

\begin{figure}[!t]
    \includegraphics[width=3.5in]{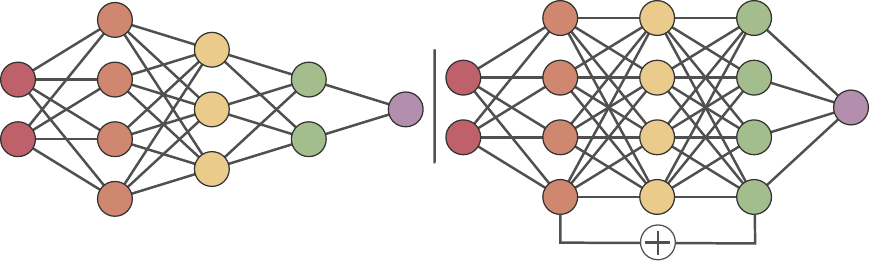}
    \caption{A plain discrete feedforward network on the left and a residual
    network on the right. The defining feature of residual networks is that the
    output from an earlier layer skips layers and is added directly to a later
    layer.}
    \label{fig:resnet}
\end{figure}

Since the success of Residual Neural Networks (ResNets)~\cite{he2016deep}, a
continuous interpretation of neural networks has gained traction. The ResNet
architecture utilizes skip connections to deal with the problem of vanishing
gradients, allowing the training of extremely deep neural networks. In a ResNet
block, the output is the sum of the usual feedforward operations on the data
and the unprocessed data entering the block. The ResNet architecture is
illustrated in Fig.~\ref{fig:resnet} on the right, with a plain feedforward
network on the left. If the hidden layers of the plain network implement a
nonlinear function $x_{k+1} = f(x_k)$, the skip connection of the ResNet
implement $x_{k+1} = x_k + f(x_k)$.  By introducing a constant $h=1$ to obtain
$x_{k+1} = x_k + hf(x_k)$, the resemblance of the skip connection to Euler's
formula for solving Ordinary Differential Equations (ODEs) is
seen~\cite{weinan2017proposal, haber2017stable, lu2018beyond}. This has lead to
the continuous-time dynamical system interpretation of ANNs.

In the continuous interpretation of ANNs, a set of layers is replaced by one
layer that parameterizes a dynamical system as a group of ODEs with state
variable $x(t)$. We assume this ODE system has the form

\begin{equation}\label{eq:generic_ode}
    x'(t) = f(x(t), u(t), \theta)
\end{equation}
in general and in our models specifically
\begin{equation}\label{eq:ode}
    x'(t) = f(Ax(t) + b) + Zu(t),
\end{equation}
where $x(t) \in \mathbb{R}^n$ is the state of the ODE at time $t$, $x'(t)$ is
the time derivative of $x(t)$, $A \in \mathbb{R}^{n \times n}$ is a weight
matrix, $b \in \mathbb{R}^n$ is a bias vector, $u(t) \in \mathbb{R}^i$ is a
time-dependent input, $Z \in \mathbb{R}^{n \times i}$ is an input matrix
(may include bias) and
$f: \mathbb{R}^n \mapsto \mathbb{R}^n$ is a vector-valued function $f(\chi) =
[f_a(\chi_1), f_a(\chi_2), \dots, f_a(\chi_n)]^T$. In Neural ODEs, the
activation function $f_a(\chi_i)$ is commonly the hyperbolic tangent, although
any differentiable activation function, or a combination of different
activation functions, can be used. Here, $\theta = (A, b, Z)$ are learned
parameters of the ODE. In feed-forward Neural ODEs there are typically no
time-dependent inputs and hence $Z=0$. On the other hand, in RNNs where $x'(t)$
is the hidden state, the input data enters the system through $Z \neq 0$. The
output of the layer is $x(t_{end})$, which is the state of the dynamical system
at the user specified final time $t_{end}$. Initial values of the ODE system
can be obtained in two ways, either as the output of the layer preceding the
ODE or set explicitly. The former is more common in feed-forward architectures
and the latter in RNNs that receive time-dependent input data. The output
$x(t_{end})$ of the ODE layer is computed using numerical methods, taking
discrete or adaptive steps with an ODE solver to solve an initial value
problem. Neural ODEs can use the adjoint method of calculating
gradients~\cite{chen2018neural} and have enabled parameterizing ODEs as several
different ANN operations, or chains of them. The primary restriction is that
the output size of the ODE layer must match the input-size of the layer.
Overall, it is possible to train a variety of ANN architectures for different
tasks as continuous networks~\cite{chen2018neural}. 

Training Neural ODEs with the adjoint method is memory efficient compared to
deep discrete networks~\cite{chen2018neural}, as the backpropagation algorithm
does not need to store intermediate ODE states to calculate gradients on the
loss function. However, the use of the adjoint method is not required for
training ODE networks. Parameterizing an ODE in place of several discrete
layers may also lead to parameter efficiency and correspondingly require less
storage space and memory, since the continuous layer can replace several
individually parameterized discrete layers. Other benefits of Neural ODEs
include using ODE solvers for speed versus accuracy tuning, and enabling ANNs
to flexibly process continuous and irregularly sampled time-series
data~\cite{chen2018neural, rubanova2019latent}. Neural ODEs have also improved
on existing ANN-based density estimation models~\cite{chen2018neural}. However,
the dynamics learned by Neural ODEs may be unnecessarily
complex~\cite{finlay2020train, kelly2020learning} and result in stiff
systems~\cite{massaroli2020stable}. Often the ODE block is the most
computationally demanding part of the network, since many numerical evaluations
of the state of the ODE are needed to obtain the output.  Hence, Neural ODEs
require more time to evaluate data, both in training and testing, compared to
traditional discrete networks~\cite{kelly2020learning}.  

It is possible to parameterize the ODE block so that the model in training has
favourable properties that facilitate learning. Antisymmetric networks are a
step toward this direction, since they guarantee that the state $x(t)$ of the
ODE system does not diverge far from or decay to the origin as $t \mapsto
\infty$~\cite{haber2017stable}. This prevents the gradient of the loss function
from vanishing or exploding and makes training the network well-posed.
Antisymmetric networks use an antisymmetric weight matrix $A = W - W^T$ that by
definition has eigenvalues $\lambda_i$ so that for all $i,
\operatorname{Re}(\lambda_i(A)) = 0$.  Furthermore, a small shift of the
eigenvalues by $\gamma$ may be applied so that $\operatorname{Re}(\lambda_i(A -
\gamma I)) = -\gamma < 0$, which improves behavior of the ODE system in the
presence of noisy data~\cite{haber2017stable}. Notice that in practical
applications $t$ is finite, hence a small $\gamma$ will not make the gradients
of the loss function vanish.  In~\cite{chang2018antisymmetricrnn} it is
demonstrated that the shifted antisymmetric weight matrix $A - \gamma I$ gives
the hidden state of recurrent neural networks a favourable property of
long-term dependence on the inputs to the system, which helps classifying data
with temporal relationships. In this work, we implement an ODE-RNN as
\begin{equation}\label{eq:antirnn}
    \begin{aligned}
        A &= W - W^T - \gamma I, \\
        x'(t) &= \text{tanh}(Ax(t) + b) + Zu(t),
    \end{aligned}
\end{equation}
where the weight matrix $A$ gives the network the desired properties that
facilitate learning and during training the parameters $W$ are learned. Other
parameters are similar to Eq.~\eqref{eq:ode}. In
Section~\ref{sec:results}, we will demonstrate compression and acceleration of
this ODE-RNN. Such networks make an interesting model compression target since
their architecture gives the system temporal memory capacity that the
compressed model must retain.

\subsection{Model Order Reduction}

A key contribution of our work is the formalization of MOR in the context of
Neural ODEs and the realization that the necessary MOR operations can be
expressed as layers of ANNs.  A powerful method for MOR of general nonlinear
systems is Proper Orthogonal Decomposition (POD)~\cite{Lumley1993} coupled with
the Discrete Empirical Interpolation Method (DEIM)~\cite{Chaturantabut2010,
barrault2004empirical}, developed in the fields of systems and control theory.
In order to compress Neural ODEs, we construct reduced order models (ROMs) of
the ODE block in trained Neural ODEs with the POD-DEIM method.   

Both POD and DEIM utilize the method of \emph{snapshots}~\cite{Sirovich87}. 
In POD, snapshots are values of the state $x_t$ of the ODE system at discrete
times $t$. A snapshot matrix $X = [x_1, x_2, \cdots, x_s]$ is collected using
numerical simulation of the ODE block with different initial values and
time-dependent input data. ANNs provide a very natural setting for gathering
snapshots, since we have access to training data that can be propagated through
the trained network and the ODE block of Neural ODEs. However, it is important
that the snapshots are collected after the model is trained, so that the
snapshots reflect true learned dynamics of the ODE layer. Moreover, the
snapshots are not used for optimization or model training. The following
explains the purpose of the snapshots.


POD approximates the original system of dimension $n$ via projection using a
low-dimensional subspace. A k-dimensional POD basis with orthonormal column
vectors $V_k \in \mathbb{R}^{n \times k}$, where $k < n$, is computed using
Singular Value Decomposition (SVD) of snapshots $V\Sigma\Psi^T =
\text{SVD}(X)$~\cite{Sirovich87}. $V_k$ is then the first $k$ left singular
vectors of the snapshot matrix, equaling the first $k$ columns of $V$.
A reduced state vector $V^T_k x(t) = \tilde{x}(t) \in \mathbb{R}^k $ is
obtained by a linear transformation and at any time $t$ an approximation of the
original, full-dimensional state vector can be computed with $x(t) \approx
V_k\tilde{x}(t)$.  Projecting the system in Eq.~\eqref{eq:ode} onto $V_k$
by Galerkin projection results in a reduced system
\begin{equation}\label{eq:pod_statespace}
  \begin{aligned}
      \tilde{x}'(t) &= V^T_kf(AV_k\tilde{x}(t) + b) + V^T_kZu(t), 
  \end{aligned}
\end{equation}
where we can  precompute the transformation of the weight and input matrices
into matrices $\tilde{A} = AV_k, \tilde{Z} = V^T_kZ$ that are used in place of
the original weight matrices in the reduced models. The reduced POD model
approximates the original system optimally in the sense that the POD subspace
has minimum snapshot reconstruction error~\cite{Lumley1993}. However, the
nonlinear form of the equation prevents computational savings as the number of
neurons has not been reduced. The size of $A$ is still $n \times k$ and $f(\cdot)$
is computed in the original dimension $n$. This is known as the lifting
bottleneck in reducing nonlinear models.

\begin{figure}[!t]
    \removelatexerror
    \begin{algorithm}[H]
    \caption{Discrete Empirical Interpolation Method}
    \label{alg:deim}
    \textbf{INPUT:}  $\{u_l\}^m_{l=1} \subset \mathbb{R}^{n}$ linearly independent \\
    \textbf{OUTPUT:} $\vec{p} = [p_1, \dots, p_m], P\in\mathbb{R}^{n\times m}$

    \begin{algorithmic}[1]
        \STATE{$p_1 = \text{argmax}(|u_1|)$} 
        \STATE{$U = [u_1], P = [e_{p_1}], \vec{p} = [p_1]$} 
        \FOR{l = 2 to m}
        \STATE{solve $(P^TU)c = P^Tu_l$ for $c$}
        \STATE{$p_l = \text{argmax}(| u_l - Uc |)$}
        \STATE{ $U \leftarrow [U \: u_l], P \leftarrow [P \: e_{p_l}],
        \vec{p} \leftarrow [\vec{p} \: p_l]$ }
        \ENDFOR
    \end{algorithmic}
\end{algorithm}
\end{figure}

Efficient evaluation of the nonlinear term can be achieved with
DEIM~\cite{barrault2004empirical, Chaturantabut2010}. DEIM extends the subspace
projection approach with an interpolation step for general nonlinear
functions. To construct an interpolated approximation of the nonlinear term, we
use
\begin{equation}
  \tilde{f}(x, t) \approx U_m(P_m^TU_m)^{-1}P_m^Tf(x, t),
\end{equation}
where the DEIM basis vectors $U_m = [u_1, u_2, \cdots, u_m] \in \mathbb{R}^{n
\times m}, \: m < n$ are the first $m$ left singular vectors of the snapshot
matrix of nonlinear vector-valued function outputs $F = [f(x_1, t_1), f(x_2,
t_2), \dots, f(x_s, t_s)]$ computed via SVD as $U_m \Sigma_U \Psi^T_U =
F$, $P_m^Tf(x, t) := f_m(x, t)$ is a nonlinear function with $m$
components chosen from $f$ according to DEIM determined interpolation
points $\vec{p} = {p_1, \cdots, p_m}$ and $P_m = [e_{p_1}, e_{p_2}, \cdots,
e_{p_m}]$ with $e_{p_i}$ being the standard basis vector $i$ of
$\mathbb{R}^n$. Together POD and DEIM form a ROM 
\begin{equation}\label{eq:deim_statespace}
  \begin{aligned}
    \tilde{x}'(t) =&
      \underbrace{V^T_kU_m(P^TU_m)^{-1}}_{N}f_m(\tilde{A}\tilde{x}(t) + b) +
      \tilde{Z}u(t),
  \end{aligned}
\end{equation}
where $N \in \mathbb{R}^{k \times m}$ can be precomputed. Now in the
prediction phase only $m$ nonlinear functions are evaluated. Due to the
structure of $f_m: \mathbb{R}^m \mapsto \mathbb{R}^m$ we can then further
compress $\tilde{A}$ so that only $m$ rows at indices $\vec{p}$ remain.
Correspondingly, we only select $m$ elements from the bias vector at indices
$\vec{p}$. Thus we have obtained the reduced order model
\begin{equation}\label{eq:rom}
  \begin{aligned}
      \tilde{x}'(t) =& Nf_m(\tilde{A}_m\tilde{x}(t) + b_m) + \tilde{Z}u(t)
  \end{aligned}
\end{equation}
where $\tilde{A}_m \in \mathbb{R}^{m \times k}$ and $b_m \in \mathbb{R}^m$.
In Eq.~\eqref{eq:deim_statespace}, the dimension $k$ of the POD
projection subspace does not need to equal the number $m$ of the DEIM
interpolation points, as $N$ can be computed even if $k\neq m$, although in
practice it is common to choose $m=k$. This means that $m, k$ can be chosen
either smaller or larger with respect to each other to reflect the complexity
of linear and nonlinear functions of the model.

Algorithm~\ref{alg:deim} shows the process of determining the DEIM
interpolation points and matrix $P$. The ordered linearly independent basis
vector set given as input is the basis $U_m$ that is obtained from snapshots of
the nonlinear function. The argmax-function returns the index of the largest
value in a given vector and $e_i$ is the $i$:th standard basis vector. For more
details see the original reference~\cite{Chaturantabut2010}. Therein an error
bound for the DEIM approximation in Euclidean space is also given. For an error
estimate of the reduced state in POD-DEIM models specifically,
see~\cite{Chaturantabut2012}.

\begin{figure}[!t]
    \includegraphics[width=3.5in]{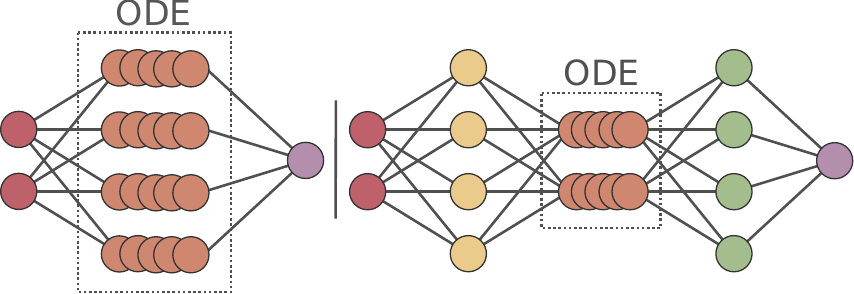}
    \caption{A Neural ODE on the left with the discretized differential
        equation block in orange color. A POD-DEIM reduced network on the
        right illustrates that the ODE block is evaluated in a low-dimensional
        subspace, with linear transformations around the ODE block. The
        networks have equal inputs and approximately equal outputs.}
        \label{fig:deimnet}
\end{figure}

An essential part of our work is building the POD-DEIM subspace projection
operations $V_k^T$, $V_k$, reduced matrices $\tilde{A}_m, \tilde{Z}$
and low-dimensional interpolation $N$ of Eq.~\eqref{eq:rom} into the Neural
ODE. Consider a
Neural ODE with an input layer, an ODE block, and a readout layer, as on the left
of Fig.~\ref{fig:deimnet}. We add two new layers, one for projecting the
ODE block into a low-dimensional subspace and one for transforming the result
of the ODE block back to the original space, as seen on the right of
Fig.~\ref{fig:deimnet}. These layers implement the computations $V_k^Tx(0)$ and
$V_k\tilde{x}(t_{end})$, respectively. The cost of these operations is offset by the
cheaper evaluation of the ODE layer, which typically makes the bulk of the
compute time of Neural ODEs. 
In the ODE layer, the reduced matrices $\tilde{A}_m, \tilde{Z}$
replace the corresponding original large weight matrices and a new
linear layer is added to implement the low-dimensional interpolation $N$.
Furthermore, if the operations around the
ODE block are linear, $V_k^T, V_k$ can be computed into those existing layers,
resulting in even cheaper online evaluation. Finally, we emphasize that
this POD-DEIM process is widely applicable to different model architectures, as
long as Eq.~\eqref{eq:generic_ode} defines the ODE block.

The DEIM method is known to exhibit lower accuracy in the presence of noise or
perturbations in the snapshots. The input function $Zu(t)$ in ODE-RNNs can
cause such non-smooth behavior and a large number of timesteps taken is
sensitive to instabilities. Hence for compressing ODE-RNNs, we use an
oversampling strategy that stabilizes the method as shown
in~\cite{peherstorfer2020stability}. In this oversampled DEIM (ODEIM), the
number of nonlinear sampling points in $\vec{p}$ becomes $m+o$ and is
decoupled from the number of basis vectors $m$ of the DEIM subspace.
The matrix $P$ then has size $n \times (m+o)$ and the matrix inverse of
Eq.~\eqref{eq:deim_statespace} is replaced with the Moore-Penrose
pseudoinverse $(P^TU_m)^{\dagger}$. In ODEIM, interpolation becomes
approximation via regression. During model evaluation, $m+o$ activation
functions are evaluated. In our results, we evaluate $m+1$ nonlinear
activation functions, which compared to vanilla DEIM has a slightly negative
effect on model acceleration but a  positive effect on accuracy of the
compressed model.

The POD-DEIM method has adjustable parameters that affect the performance of the
reduced model. The dimension $k$ of the linear projection matrix and the number of
interpolation points $m$ are typically set according to the decay of the
singular values of the snapshot matrices $X, F$, but both can be adjusted
independently. The number of snapshots can be controlled in two ways, the
number of samples of training data to use and the number of snapshots of model
dynamics to save per sample. Computing the SVD of the snapshot matrix is the
most computationally demanding step of the MOR pipeline, with memory being the
typical bottleneck.

\subsection{Benchmark compression methods}

We compare the performance of the POD-DEIM method to two other ANN compression
methods presented in the literature. To the best of our knowledge, no other
studies using these methods to compress Neural ODEs have been published before.

The first method compresses the weight matrix of the ODE block into two
low-rank matrices~\cite{denton2014exploiting, girshick2015fast} using SVD based
truncation. This reduces the overall number of multiply-adds needed to evaluate
the layer but, unlike POD-DEIM, does not change the number of
activation functions. Given a fully connected layer with activation $f(Ax + b)$
and $A \in \mathbb{R}^{n \times l}$, we apply SVD to the weight matrix to
obtain $\Phi \Sigma \Psi^T = A$. Then, we only keep the first $k < n$ singular
values by truncation so that $\Sigma_k = \text{diag}(\sigma_1, \dots \sigma_k)
\in \mathbb{R}^{k \times k}$ and the first $k$ left and right singular vectors
so that $\Phi_k = \{\phi_1, \dots, \phi_k\} \in \mathbb{R}^{n \times k}, \Psi_k
= \{\psi_1, \dots, \psi_k\} \in \mathbb{R}^{l \times k}$. An approximation to A
is then obtained as $\tilde{A} = \Phi_k\Sigma_k \Psi_k^T$. In ANNs we can
implement the approximation in two consequent fully connected layers that only
use activation and bias after the last layer.  The connection weights of the
first layer are $\Sigma_k \Psi^T_k \in \mathbb{R}^{k \times l}$ and the second
weights are $\Phi_k$, which changes the parameter count from $nl$ to $k(n+l)$.
Notice that in a single-layer ODE block we have $l = n$. This method is
theoretically sound for compressing the ODE block. It retains the most
parameters out of the compression methods we test, meaning that it is expected
to yield low to moderate acceleration. Since the weight
truncation only uses the learned weights as inputs, the method does not depend
on any other data, unlike POD-DEIM which requires gathering snapshots of the
ODE dynamics.

The second method we used for comparison is based on computing an importance
score for neurons, called average percent of zeros (APoZ) method,
following~\cite{hu2016network}. There it is suggested to prune neurons that
have the largest number of zero activation values over the training or testing
data set. The corresponding rows of the weight, bias and input matrices are
then removed altogether. Moreover, with Neural ODEs specifically, for each
removed row of the weight matrix we also remove the corresponding column. The
importance score of the $c$:th neuron in a layer is defined as 
\begin{equation}
    \text{APoZ}_c = \frac{\sum^N_k\sum^M_jf(O_{c,j}(k) = 0)}{NM},
\end{equation}
where $N$ is the number of examples used to calculate the score and $M$ is the
dimension of the output feature map $O$. In Neural ODEs it is common to use
hyperbolic tangent activations, hence we adapt the scoring so that neurons with
lowest absolute activation magnitudes are eliminated. We compute the score
after the last timestep of the ODE block, as this is the value that gets
propagated to the following layers in the network. We only use training data
for computing the score. This method does not account for the dynamics of the
ODE block, as scoring is computed only at the last timestep of the ODE block,
hence we evaluate the performance in detail here. Overall, this method applies
the largest amount of compression out of the methods we test here.

\section{Results}\label{sec:results}

We implemented two Neural ODE models and trained them on different
classification tasks; one Neural ODE with a convolutional ODE block to classify
digits of the MNIST dataset and one ODE-RNN to classify digits in the
pixel-MNIST task, where the network is given one pixel at a time as input. We
trained both models using data-augmentation with random rotations and affine
translations in order to prevent overfitting, using a decaying learning
    rate starting from $0.04$, stochastic gradient descent optimizer and
    cross-entropy loss function on image labels. After training, the POD-DEIM
    snapshots were collected by feeding each image in the training data to the
    network, and saving the ODE state and activation function values at every
    second timestep. POD and DEIM bases were computed using a memory efficient
partitioning strategy~\cite{liang2016split}. Next, we compressed the models
using POD-DEIM, APoZ trimming and weight truncation and evaluated their
performance directly after compression, after 3 epochs of fine-tuning and after
excessive fine-tuning on training data. The performance of each compressed
model was compared to the respective original model using Top-1 accuracy, Top-3
accuracy and wall-time as metrics. We restrict fine-tuning to the layers
following the ODE block in order to get a better view of how well the
compressed blocks maintain their computational capacity. Our results below have
been implemented using the PyTorch machine learning
framework~\cite{pytorch2019}. To implement Neural ODEs, we additionally used
the TorchDiffEq~\cite{chen2018neural} and Torchdyn~\cite{poli2020torchdyn}
python packages.  Execution times are true wall-times realized on Intel Xeon
E5-2680 v3 2.5GHz processors and measured as the time it takes to classify
every item in the test dataset. For the convolutional network, we use a
single core, and for the recurrent network four processor cores.

In Table~\ref{table:theoretical}, we show the number of parameters in a
theoretical Neural ODE and after it has been compressed with each compression
method. We consider a nonlinear ODE block of the form $x'(t) = f(Ax(t))$ where
the weight matrix has dimensions $A \in \mathbb{R}^{n \times n}$, hence there
are $n$ activation functions. The ODE block is preceded by a layer with weight
matrix $W_1 \in \mathbb{R}^{n \times i}$ and followed by a layer with weight
matrix $W_2 \in \mathbb{R}^{o \times n}$. In Table~\ref{table:theoretical}, $k$
denotes the dimension of the compressed ODE block and $n \times i, o \times n$
are the number of weights in the layers preceding and following the ODE block,
assuming those layers exist.  With regards to the model speed, the
complexity of the
ODE layer is significantly more important than the sizes of the surrounding
layers. This is because the ODE function is evaluated multiple times; at least
once for each intermediate time point in $[t_{0}, t_{end}]$. The
architecture of the original ANN affects the achieved compression. The POD-DEIM
and APoZ methods can reduce the ODE layer to a dimension that is independent of
the original size $n$, although the interpolation in POD-DEIM requires an
additional $k^2$ operations compared to APoZ. Moreover, for POD-DEIM the best
case is realized when the preceding and following operations are linear,
whereas the APoZ method reaches the best case even if they are nonlinear. With
the APoZ method, some nonlinear operations, such as max pooling, require that
we maintain a lookup table of pooled indices or insert the $k$ compressed
values to their respective indices in an $n$ size vector, which is the worst
case result for the method. SVD truncation is not affected by the surrounding
layers. 
\begin{table}[!t]
    \begin{threeparttable}
    \renewcommand{\arraystretch}{1.2}
    \caption{Theoretical layer sizes}\label{table:theoretical}
    \begin{tabular}{lp{5mm}p{10mm}p{12mm}p{12mm}}
    Model                 & ODE weights & ODE activations & Preceding weights  & Following weights  \\ \hline
    Original model        & $n^2$       &    $ n $        & $ni$             & $on$  \\
    POD-DEIM, best case   & $2k^2$       &    $ k $        & $ki$             & $ok$  \\
    POD-DEIM, worst case  & $2k^2$       &    $ k $        & $n(i + k)$       & $n(o+k)$    \\
    APoZ, best case       & $k^2$       &    $ k $        & $ki$             & $ok$ \\
    APoZ, worst case      & $k^2$       &    $ k $        & $ki$             & $n(o + 1)$ \\
    SVD                   & $2kn$       &    $ n $        & $ni$             & $on$
    \end{tabular}
    \begin{tablenotes}
        \small
        \item $n$: ODE block dimension, $k$: compressed ODE dimension, $i$: neurons
            in the layer preceding the ODE block, $o$: neurons in the layer
            following the ODE block.
    \end{tablenotes}
    \end{threeparttable}
\end{table}

\begin{figure*}[!t]
    \includegraphics[width=7in]{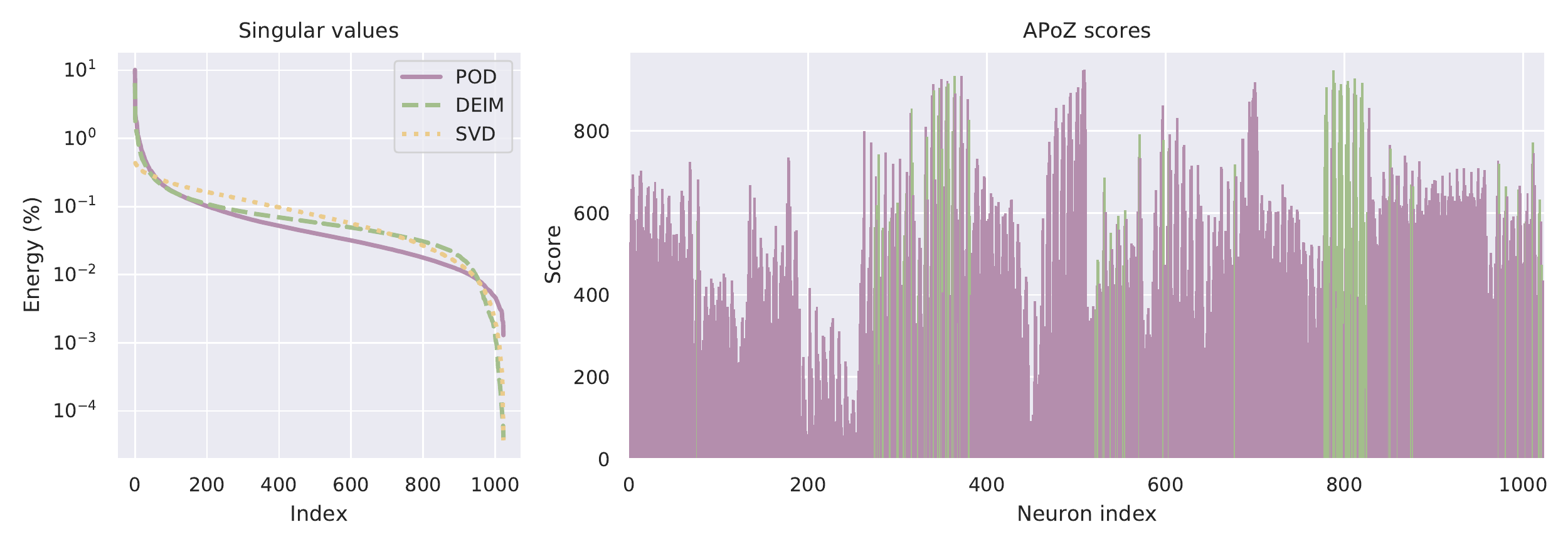}
    \caption{Metrics for the convolutional Neural ODE. Left: Singular
    values POD (purple) and DEIM (green) snapshot matrices and singular values of
    the weight matrix of the ODE block (yellow). Right: histogram of APoZ
    scores, with the first 100 DEIM indices indicated in green color.}
    \label{fig:conv_data}
\end{figure*}

\subsection{Convolutional Neural ODE}

We illustrate the performance of reduced models on the MNIST dataset of
handwritten digits~\cite{lecun1998mnist}. For our first task we train a
convolutional Neural ODE that uses 16 convolutional kernels and
\textit{tanh}-activation in the ODE block, with $Z=0$ (see Eq.~\eqref{eq:ode}.
Before the ODE block we use a convolutional layer with $3 \times 3$ kernels and
rectified linear unit (ReLU) activations that outputs 16 channels into $3
\times 3$ max pooling. After the ODE block we use $3 \times 3$ max pooling into
a linear readout layer with ReLU activations. The ODE block propagates the data
for $t = [0, 1]$ with timestep $dt=0.1$ using a fourth order fixed-step
Runge-Kutta solver. The initial value of the ODE is the output from the prior
layer. We use the adjoint method for training the ODE
weights~\cite{chen2018neural}. In order to implement model order reduction and
model compression of the convolutional ODE block, the convolution operator is
needed in matrix form. After training, we replaced the original convolution
operator with an equivalent operation implemented as a linear layer. The
conversion yields a sparse Toeplitz matrix. With this operation, 16 convolution
channels correspond to 1024 neurons and activation functions. All reported wall
times are obtained with this conversion and compressed models are compared to
these times. The reported wall times are medians of 10 consecutive evaluations
of the entire validation data set. The trained network achieves 97.9\% top-1
and 99.7\% top-3 accuracies on held-out test dataset of 10 000 MNIST images and
it takes 9.4 seconds to run the model on the test dataset on CPU.

\begin{figure*}[!t]
    \includegraphics[width=\textwidth]{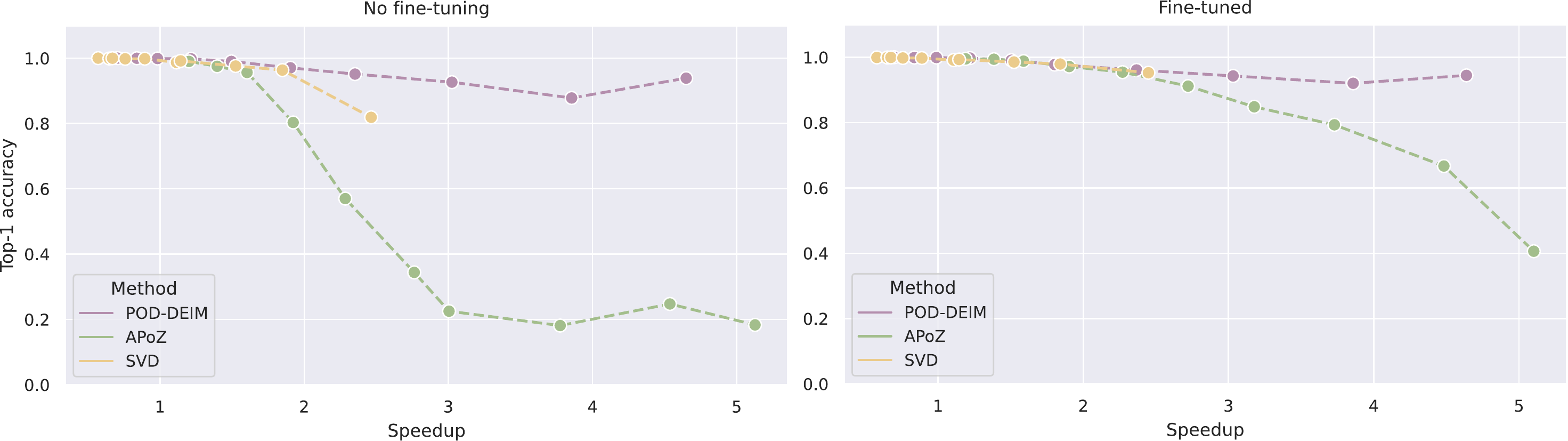}
    \caption{Performance of reduced convolutional Neural ODEs on the MNIST
        dataset, relative to the original model. Left plot shows results
        without fine-tuning, and right plot with three epochs of tuning. The
        x-axis shows achieved acceleration, while the y-axis shows reduced
        model top-1 accuracy divided by full model accuracy. Reduced model
    dimension decreases to the direction of improving speedup. Absolute
performance can be seen in \tableconv{}.}
    \label{fig:mnist_conv_cpu}
\end{figure*}

In order to determine the suitability of our chosen compression methods, we
looked at several metrics of the trained Neural ODEs, namely decay of singular
values in the POD-DEIM snapshot matrices $X, F$ and the weight matrix $A$. 
Moreover, we looked at the distribution of APoZ scores and their relation to
DEIM interpolation indices. Fig.~\ref{fig:conv_data} shows these metrics for
the convolutional Neural ODE. The left plot shows sorted singular values of the
solution snapshot matrix $X$ (POD) and $F$ (DEIM) of the ODE block, gathered
from training data, in purple and green colors, respectively. The logarithmic
y-axis shows the magnitude of a given singular value divided by the sum of all
singular values. Both POD and DEIM singular values collapse rapidly, indicating
that a small number of singular vectors can span a space where the majority of
dynamics are found. This is an important indicator for the suitability of the
POD-DEIM method. In the same plot, the yellow plot shows the singular values of
the weight matrix $A$ of the ODE block. These values are useful for determining
a rank for the SVD based truncation method that we compare against POD-DEIM.
Singular values of $A$ also decay exponentially, meaning that a good low-rank
approximation to $A$ can likely be found. The right plot of
Fig.~\ref{fig:conv_data} shows APoZ scores for each neuron in the ODE block,
with low-scoring neurons being the least important. Additionally, we have
indicated the first 100 interpolation points $p_i$ chosen by the DEIM method in
green color. All neurons seem to contribute to the model, with only a small
number of relatively low-scoring neurons seen, which could make the APoZ method
challenging to apply in the Neural ODE context. Some consensus between the
methods is seen, as neurons with low APoZ scores are not in the top DEIM
selections either.

We wanted to find out how the POD-DEIM method compares to existing model
compression methods when compressing only the ODE block of a convolutional
network. For POD-DEIM dimensions, we have chosen $k=m$ and use dimension $k$
compressed models, which corresponds in number of activation functions to
dimension $k$ APoZ trimmed models. The results on the MNIST dataset are seen in
Fig.~\ref{fig:mnist_conv_cpu}, 
where y-axis, model accuracy, is computed as reduced model top-1 accuracy divided by
original model result and x-axis, achieved speedup, as original model test time
divided by reduced model test time. Each dot represents a compressed model
dimension, which decreases to the direction of increased speedup. While the
dimensions are not comparable between methods, the graphs indicate how much
accuracy is maintained at a given acceleration amount. \tableconv{} presents
the absolute performance values of compressed models without fine-tuning, with
three epochs of fine-tuning after model compression and with retraining (30
epochs of training after compression).

Fig.~\ref{fig:mnist_conv_cpu} shows that the POD-DEIM method retains the
highest accuracy when comparing to similar speedups from other methods. Our
POD-DEIM reaches over four fold speedup in runtime in this task. The SVD
method is also accurate, but cannot reach speedups greater than two times
the original runtime, explained by the method having to evaluate all the
original nonlinear activation functions. The APoZ method reaches the
fastest run times but loses the most amount of accuracy. After three epochs
of fine-tuning, the APoZ method improves in accuracy, but not to the level
of performance that the other methods show even without tuning. Overall,
the POD-DEIM method achieves the best profile in terms of accuracy retained
for speedup gained. A big contributor is the interpolation matrix $N$,
which makes the method slower compared to APoZ, but the cost is justified
by the better accuracy. When looking at absolute performance values,
\tableconv{} shows that at dimension 350, the POD-DEIM method is two times
faster to run than the original model, while achieving 93.1\% top-1
accuracy. At dimension 50, the POD-DEIM model is over four times faster
than the original model, still reaching 91.1\% top-1 accuracy without
fine-tuning.

\subsection{Recurrent Neural ODE}

For our second study, we designed a task in which Neural ODEs have an advantage
over discrete networks. ODE-RNNs~\cite{rubanova2019latent} and antisymmetric
RNNs~\cite{haber2017stable, chang2018antisymmetricrnn} are one such example.
Using a continuous hidden state, ODE-RNNs enable flexible inference on
continuous-time data even with irregularly sampled or missing values and
antisymmetric models have spectral properties that are favourable for RNNs, as
described in Section~\ref{sec:methods}. We implemented a shifted antisymmetric
ODE-RNN for the pixel MNIST task. In this task, the network sees a pixel of a
handwritten digit at a time and after iterating over each pixel outputs the
class of the image.  The length of the input sequence is 784 steps,
corresponding to a flattened MNIST image, and the network sees each pixel for
$dt = 0.1$ seconds. Following~\cite{chang2018antisymmetricrnn}, we initialize
the ODE weights from zero mean, unit variance normal distribution as well as
train and test the network with the Euler discretization method. We use 512
hidden units with a hyperbolic tangent nonlinearity and the shifted
antisymmetric weight matrix as shown in Eq.~\ref{eq:antirnn}. Initial value of
the ODE is set to $x(0) = 0$. The ODE
block is followed by a linear readout layer. Compared to the convolutional
network, this model is smaller in neuron count but needs to incorporate the
time-dependent input data.  The trained network achieves 96.2\% accuracy on
held-out MNIST test data in 48.1 seconds.

\begin{figure*}[!t]
    \includegraphics[width=\textwidth]{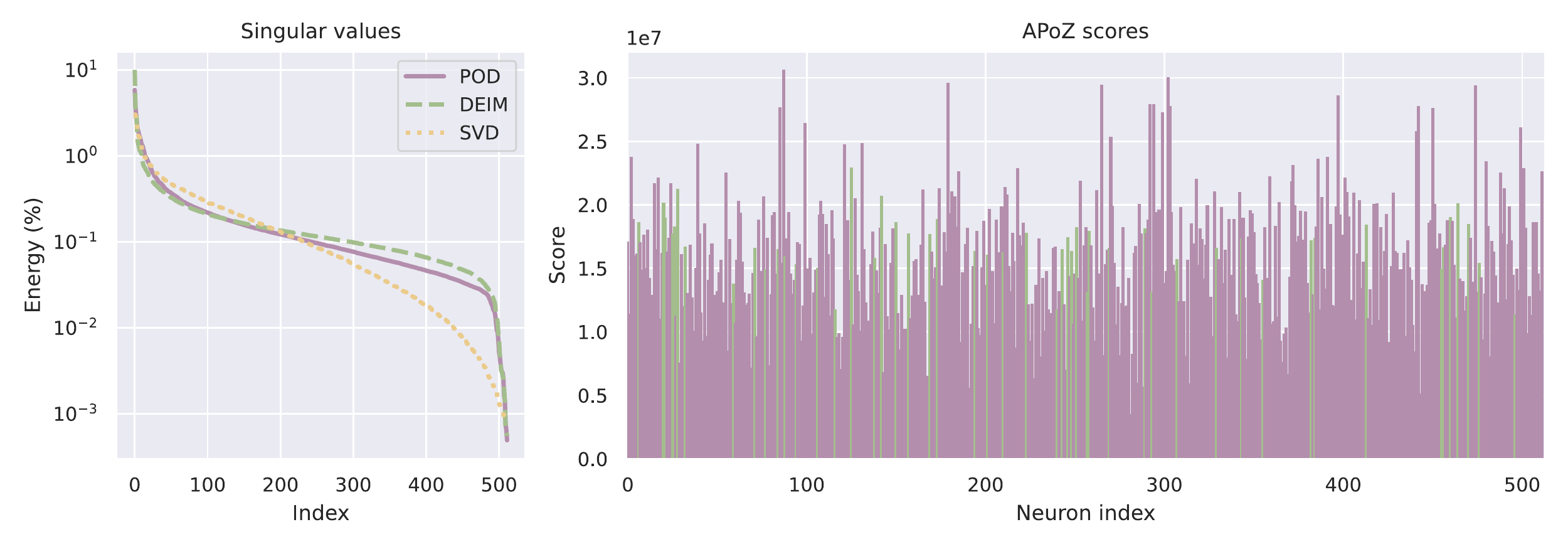}
    \caption{Metrics of the ODE-RNN. Left: Singular values of POD (purple) and
    DEIM (green) snapshot matrices and singular values of the weight matrix of
    the ODE block (yellow). Right: histogram of APoZ scores, with the first 50
    DEIM indices indicated in green color.}
    \label{fig:rnn_data}
\end{figure*}

Fig.~\ref{fig:rnn_data} shows compression metrics for the ODE-RNN. The left
plot shows sorted relative singular value magnitude of the solution snapshot
matrices $X$ (POD) and $F$ (DEIM) and the weight matrix $A$ of the ODE block in
purple, green and yellow, respectively. For this model, the singular values of
the snapshot matrices as well as the weight matrix also decay rapidly. However,
with regards to POD-DEIM, the singular values of the snapshot matrices do not
decay as fast as those obtained from the convolutional model and there is a
considerable amount of total energy contained in the singular values until rank
500. This indicates that the dynamics of the ODE block are more difficult to
approximate in low dimensions than those learned by the convolutional model.
Such a result is expected, because the ODE-RNN has a time-dependent input
contributing to the hidden state and the forward propagations are computed for
a longer time span than in our earlier example, allowing for richer dynamics.
On the other hand, singular values of the shifted antisymmetric weight matrix
decay more rapidly than in the earlier task, indicating suitability of
SVD-based truncation of weights. The APoZ scores on the right of
Fig.~\ref{fig:rnn_data} show the same pattern as earlier, where it is not
straightforward to detect the most important neurons conclusively. This is
expected, given that in the context of ODE blocks, the output activity at last
time step is not a conclusive measure of neuron importance. 

\begin{figure*}[!t]
    \includegraphics[width=7in]{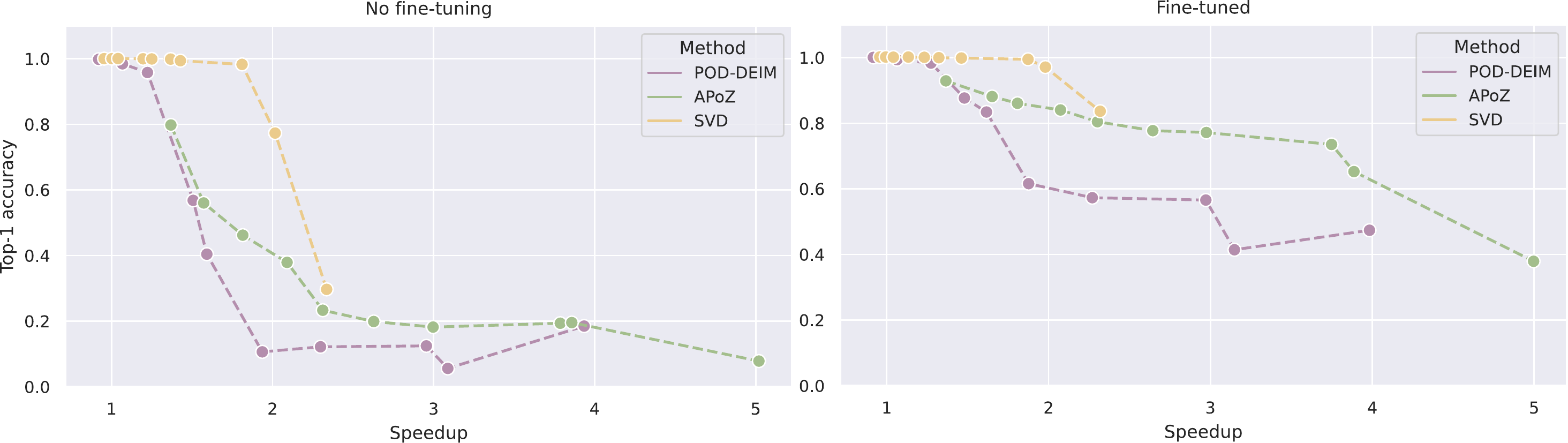}
    \caption{Relative performance of reduced ODE-RNNs on the pixel MNIST task.
        Left plot shows results without fine-tuning, and right plot with three
        epochs of tuning. The x-axis shows achieved acceleration, while the
        y-axis shows reduced model top-1 accuracy divided by full model
        accuracy. Reduced model dimension decreases to the direction of
    improving speedup. Absolute values can be seen in \tablernn{}.}
    \label{fig:antisymm_cpu}
\end{figure*}

We then evaluated the selected compression methods on the pixel MNIST task
using our ODE-RNN. The results relative to the performance of the original
model are seen in Fig.~\ref{fig:antisymm_cpu}. Absolute results of compressed
models without fine-tuning, with three epochs of fine-tuning and with
exhaustive tuning in \tablernn{}. 
This task is more challenging for the reduction methods than the
convolutional network. Possible factors are the dense structured antisymmetric
weight matrix or the time-dependent input. Here, the SVD truncation method
maintains the most accuracy. As predicted by the singular values of the
snapshot matrices, the overall performance of the POD-DEIM method on this model
is not as satisfying as with the convolutional model. On this model the
POD-DEIM method is faster than the original model until dimension 425, at which
point the accuracy is 94.6\%, while the APoZ method remains faster at all
times. With regards to the trade-off between run time and accuracy, the SVD
method maintains the most accuracy for speed gained. POD-DEIM shows good
accuracy at high dimensions, although accuracy drops rapidly so that the
overall performance is level with the APoZ method. With exhaustive fine-tuning
the APoZ method may be fit for compressing the ODE-RNN to a low dimension,
since APoZ yields the best absolute acceleration. The APoZ compressed model
also seems easier to fine-tune than the POD-DEIM version, since more accuracy
is recovered in three fine-tuning epochs.

\section{Discussion}\label{sec:discussion}

In this work we have studied compression of artificial neural networks that
contain ordinary differential equations as layers (Neural ODEs). After training
and compressing the networks, we evaluated their accuracies as well as
execution speeds. Although in the literature many compression methods are
applied in the training loop, the training of Neural ODEs is slower than
discrete networks to begin with, making it attractive to compress the network
in a separate step after training. We also assessed fine-tuning of the
compressed network to see how much accuracy can be recovered. We focused on the
runtime and classification accuracy of compressed models, since Neural
ODEs are already very memory and parameter efficient by design. 

Here we showed how to formulate the Proper Orthogonal Decomposition with
Discrete Empirical Interpolation Method (POD-DEIM) into compressing Neural
ODEs. This method has been developed for accurately approximating dynamical
systems in low dimensions and the continuous-time network formalism of Neural
ODEs makes it possible to use POD-DEIM in deep learning applications.
Additionally, we compared the POD-DEIM and APoZ methods to Singular Value
Decomposition (SVD) based truncation. The performance was measured with two
different architectures, a convolutional Neural ODE and a recurrent ODE
architecture, which were trained on the MNIST digit classification task.  

We hypothesized that some existing neuron pruning methods that measure
neuron importance based on the final output of the layer only, such as the
average percent of zeros (APoZ) approach, are not optimal for compressing
Neural ODEs. The reason is that these methods do not account for the dynamics
of the ODE block when identifying pruning targets. Based on our results,
the proposed approach for Neural ODE compression depends on the model
architecture. We found the POD-DEIM method to achieve the most favourable
continuum between accuracy and speedup on compressing the convolutional
architecture, as seen in Fig.~\ref{fig:mnist_conv_cpu}. The SVD-based
weight truncation is the simplest of the studied methods and it performed
well on both tasks, although it yielded less absolute acceleration than the
other methods. The ODE-RNN was challenging to accelerate, and here the
weight truncation method is our favoured choice based on
Fig.~\ref{fig:antisymm_cpu}. In this task, the APoZ method performed better
or equally to POD-DEIM. While the POD-DEIM method is based on approximating
the time-trajectory in a low-dimensional subspace, there are also MOR
methods, such as the Balanced Truncation, that are designed to approximate
input-output behavior of high-dimensional systems. Such a method could be a
good tool for compressing RNN models with time-dependent input, once the
applicability of these methods to nonlinear systems develops
further~\cite{benner2017balanced}.

In this study we only fine-tuned the layers following the ODE block, keeping
results more indicative of raw performance of the methods. If the entire model
were tuned, POD-DEIM and SVD truncation have more parameters available for
fine-tuning than the APoZ method, allowing for greater accuracy. On the other
hand, the small number of parameters is what makes APoZ compressed models the
fastest out of the three methods, hence it should be favored if there are no
restrictions on fine-tuning and retraining time. Moreover, compared to
POD-DEIM, the APoZ method is more data and memory efficient, since it does not
need snapshots of trajectories of ODE dynamics nor does it compute large matrix
decompositions. Overall, we found that POD-DEIM can achieve good acceleration
and accuracy of compressed models, it has a solid theoretical foundation and
several extensions, while also being the most adjustable and optimizable
approach of the methods used in this study.

We expect our results with the POD-DEIM method motivate machine learning
researchers to look into the possible applications of other MOR methods
developed in control theory and related fields. For example, we assessed here
only the original DEIM~\cite{Chaturantabut2010} and the
ODEIM~\cite{peherstorfer2020stability} methods, although in recent years many
advanced versions have been developed.  These include novel interpolation point
selection methods~\cite{drmac2016new} and methods with adaptive basis and
interpolation point selection capabilities~\cite{LDEIM, ADEIM,
chaturantabut2017temporal} that allow the model to change its dimension or
projection subspace online. It is worth noting that the POD-DEIM method aims at
approximating the dynamics of the original system and is not aware of
classification or other performance indicators of the Neural ODE. A potential
future development could explore connecting the low-dimensional dynamics to
downstream network performance.

A limitation of the present work is that we could only study relatively small
models, due to known challenges with training very large Neural
ODEs~\cite{finlay2020train}. In small models the expected gains from
compression methods are lesser than in large models. Moreover, we only
discussed computation times on CPUs, as on GPUs the differences between
original and compressed models were minor due to the small size of the models.
While the computational speedup from POD-DEIM needs to be verified on GPUs and
extremely large models, theoretical reductions in parameter and activation
function count, comparable results from existing literature as well as our
results on the CPU indicate that the method should perform equally well.

Our results were obtained with fixed-step ODE solvers. We found that
compressing Neural ODEs sometimes induces stiffness to the low-rank models
(data not shown). This could cause adaptive ODE solvers to be slower,
mitigating some of the speedup from compression. However, a systematic study of
this behavior was not sought here and fixed-step ODE solvers were chosen to
quantitatively compare the methods across dimensions. A future study is needed
to quantify the effects of stiffness that may rise when compressing Neural
ODEs.

\section{Conclusion}\label{sec:conclusion}

In this study, we addressed speeding up and compressing artificial neural
networks with continuous layers. Our POD-DEIM method, which has not previously
been used to compress Neural ODEs, provides attractive results for
accelerating convolutional Neural ODEs and equal results to neural pruning
methods for accelerating recurrent Neural ODEs. POD-DEIM properly considers
the trajectory of the ODE layer in an efficient manner. Additionally,
POD-DEIM has adjustable parameters and advanced variations that can be used
to optimize the speed versus accuracy trade-off for each model.

We expect our acceleration method to be useful when Neural ODEs are deployed on
low-power hardware or battery-powered devices, for example wearable devices,
medical monitoring instruments or smart phones. The dynamical system formalism
for neural networks also allows other model order reduction methods to be
applied in deep learning, which will provide for interesting future studies.

\section*{Acknowledgment}

This research has received funding from the Academy of Finland (decision no.
297893 M.-L.L. and decision no. 310489 L.P.) and the European Union’s Horizon
2020 Framework Programme for Research and Innovation  under the Specific Grant
Agreement Nos. 785907 (Human Brain Project SGA2) and 945539 (Human Brain
Project SGA3). M.L. acknowledges the support from Tampere University Graduate
School and The Finnish Foundation for Technology Promotion.

Neural network illustrations generated partly with
\url{http://alexlenail.me/NN-SVG/index.html}

\bibliographystyle{IEEEtran}
\bibliography{refs}  

\begin{IEEEbiography}[{\includegraphics[width=1in, height=1.25in,
    keepaspectratio]{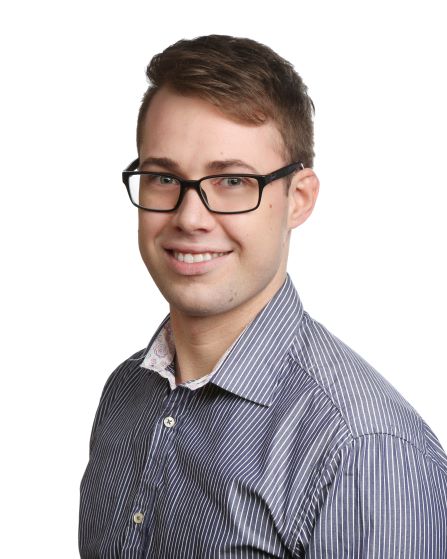}}]{Mikko Lehtimäki}
Biography text here.
\end{IEEEbiography}

\begin{IEEEbiography}{Lassi Paunonen}
Biography text here.
\end{IEEEbiography}

\begin{IEEEbiography}{Marja-Leena Linne}
Biography text here.
\end{IEEEbiography}

\newpage
\appendix[Performance of compressed Neural ODEs]

\begin{table}[h]
    \tablesize
    \begin{threeparttable}
    \caption{Absolute Performance of Reduced Convolutional Neural ODEs.}
    \label{table:conv}
    \begin{tabular}{ll|l|ll|l|ll|l|l}
    Dim.  & \multicolumn{3}{c}{Top-1 (\%)} & \multicolumn{3}{c}{Top-3 (\%)}& \multicolumn{3}{c}{Runtime (s)} \\ \hline
    \multicolumn{10}{c}{Original Model} \\ 
    1024  & \multicolumn{3}{l}{\convto} & \multicolumn{3}{l}{\convtt}  & \multicolumn{3}{l}{\convruntime}     \\
    \multicolumn{10}{c}{DEIM} \\ 
    50  & 91.9 & 92.5 & 92.9 & 98.6 & 98.9 & 99.0 & 2.4  & 2.4  & 2.0 \\
    150 & 85.9 & 90.1 & 90.9 & 97.2 & 98.4 & 98.7 & 2.9  & 2.9  & 2.4  \\
    250 & 90.7 & 92.3 & 92.8 & 98.2 & 98.8 & 99.0 & 3.7  & 3.7  & 3.1  \\
    350 & 93.1 & 94.1 & 94.4 & 99.0 & 99.1 & 99.2 & 4.8  & 4.7  & 4.0  \\
    450 & 95.0 & 95.7 & 95.9 & 99.3 & 99.5 & 99.5 & 5.9  & 6.2  & 5.3  \\
    550 & 96.9 & 97.0 & 96.9 & 99.7 & 99.7 & 99.6 & 7.5  & 7.4  & 6.3  \\
    650 & 97.7 & 97.6 & 97.6 & 99.7 & 99.7 & 99.6 & 9.2  & 9.1  & 10.1  \\
    750 & 97.8 & 97.8 & 97.8 & 99.7 & 99.8 & 99.7 & 11.4 & 11.3 & 13.4  \\
    850 & 97.8 & 97.8 & 97.9 & 99.8 & 99.7 & 99.7 & 13.3 & 13.3 & 16.8  \\
    950 & 97.9 & 97.9 & 97.9 & 99.7 & 99.7 & 99.7 & 15.8 & 16.0 & 19.9  \\
    \multicolumn{10}{c}{APoZ} \\ 
    50  & 17.9 & 39.8 & 43.6 & 31.2 & 75.5 & 79.9 & 2.2 & 2.2 & 2.2 \\
    150 & 24.2 & 65.3 & 68.7 & 48.8 & 88.5 & 90.9 & 2.5 & 2.5 & 2.5  \\
    250 & 17.7 & 77.6 & 81.3 & 76.3 & 95.0 & 95.6 & 3.0 & 3.0 & 2.9  \\
    350 & 22.0 & 83.0 & 85.4 & 83.4 & 96.6 & 97.2 & 3.7 & 3.5 & 3.1  \\
    450 & 33.7 & 89.2 & 91.2 & 84.2 & 98.4 & 98.5 & 4.0 & 4.1 & 3.6  \\
    550 & 55.8 & 93.4 & 94.5 & 97.1 & 99.2 & 99.2 & 4.9 & 4.9 & 4.2  \\
    650 & 78.6 & 95.1 & 95.7 & 97.7 & 99.5 & 99.4 & 5.8 & 5.9 & 5.8  \\
    750 & 93.6 & 96.7 & 97.0 & 99.2 & 99.6 & 99.6 & 7.0 & 7.0 & 7.1  \\
    850 & 95.4 & 97.3 & 97.3 & 99.4 & 99.6 & 99.6 & 8.0 & 8.1 & 8.8  \\
    950 & 96.9 & 97.4 & 97.5 & 99.6 & 99.7 & 99.6 & 9.3 & 9.4 & 10.4  \\
    \multicolumn{10}{c}{SVD} \\ 
    50  & 80.1 & 93.2 & 93.7 & 97.9 & 99.2 & 99.3 & 4.5 & 4.6 & 4.6 \\
    150 & 94.3 & 95.9 & 96.1 & 99.2 & 99.5 & 99.4 & 6.1 & 6.1 & 5.3  \\
    250 & 95.5 & 96.5 & 96.5 & 99.3 & 99.7 & 99.6 & 7.3 & 7.3 & 6.5  \\
    350 & 96.6 & 97.0 & 97.1 & 99.6 & 99.7 & 99.6 & 10.0 & 10.1  & 9.0  \\
    450 & 97.0 & 97.2 & 97.3 & 99.7 & 99.7 & 99.6 & 9.8 & 9.7 & 8.7  \\
    550 & 97.7 & 97.6 & 97.7 & 99.7 & 99.8 & 99.7 & 12.5 & 12.6 & 11.2  \\
    650 & 97.7 & 97.7 & 97.7 & 99.7 & 99.8 & 99.6 & 14.7 & 14.7 & 13.3  \\
    750 & 97.8 & 97.8 & 97.9 & 99.7 & 99.7 & 99.7 & 17.2 & 17.1 & 17.4  \\
    850 & 97.9 & 97.8 & 97.9 & 99.7 & 99.8 & 99.7 & 16.7 & 16.5 & 16.9  \\
    950 & 97.9 & 97.8 & 97.8 & 99.7 & 99.8 & 99.7 & 19.6 & 19.3 & 20.1  \\
    \end{tabular}
    \begin{tablenotes}
        \small
        \item Each column of Top-1, Top-3 and Runtime shows the results for the
            models directly after reduction, after three fine-tuning epochs and
            after retraining (30 epochs), in that order, separated by vertical lines.
    \end{tablenotes}
    \end{threeparttable}
\end{table}

\begin{table}[h]
    \tablesize
    \begin{threeparttable}
    \caption{Absolute performance of reduced ODE-RNNs.}
    \label{table:rnn}
    \begin{tabular}{ll|l|ll|l|ll|l|l}
    Dim.  & \multicolumn{3}{c}{Top-1 (\%)} & \multicolumn{3}{c}{Top-3 (\%)}& \multicolumn{3}{c}{Runtime (s)} \\ \hline
    \multicolumn{10}{c}{Original Model} \\ 
    512  & \multicolumn{3}{l}{\rnnto} & \multicolumn{3}{l}{\rnntt}  & \multicolumn{3}{l}{\rnnruntime}     \\
    \multicolumn{10}{c}{DEIM} \\ 
    25  & 17.8 & 45.5 & 47.2 & 38.5 & 74.7 & 76.0 & 12.3 & 12.1 & 12.4 \\
    75  & 5.4  & 39.8 & 48.7 & 24.9 & 65.3 & 69.8 & 15.6 & 15.3 & 15.8 \\
    125 & 12.0 & 54.4 & 61.5 & 33.2 & 81.8 & 86.4 & 16.3 & 16.2 & 16.3 \\
    175 & 11.7 & 55.1 & 61.6 & 33.4 & 83.2 & 87.5 & 21.0 & 21.2 & 21.1 \\
    225 & 10.2 & 59.2 & 62.7 & 32.4 & 86.3 & 88.6 & 24.9 & 25.7 & 26.2 \\
    275 & 38.9 & 80.2 & 85.1 & 71.6 & 96.0 & 97.4 & 30.3 & 29.8 & 30.6 \\
    325 & 54.6 & 84.3 & 87.7 & 83.9 & 97.0 & 98.0 & 32.0 & 32.5 & 32.4 \\
    375 & 92.1 & 94.5 & 95.2 & 99.2 & 99.4 & 99.4 & 39.4 & 37.8 & 40.5 \\
    425 & 94.6 & 95.5 & 95.8 & 99.4 & 99.5 & 99.5 & 45.2 & 45.3 & 48.5 \\
    475 & 96.0 & 96.2 & 96.2 & 99.6 & 99.6 & 99.6 & 52.4 & 52.5 & 52.8 \\
    \multicolumn{10}{c}{APoZ} \\ 
    25  & 7.5  & 36.5 & 40.2 & 25.9 & 69.1 & 72.2 & 9.6  & 9.6  & 9.3 \\
    75  & 18.8 & 62.7 & 69.4 & 42.2 & 87.9 & 91.0 & 12.5 & 12.4 & 12.2 \\
    125 & 18.6 & 70.7 & 76.1 & 44.0 & 92.5 & 94.9 & 12.7 & 12.8 & 12.7 \\
    175 & 17.5 & 74.2 & 80.2 & 46.6 & 93.9 & 96.1 & 16.1 & 16.2 & 16.0 \\
    225 & 19.1 & 74.7 & 81.1 & 43.7 & 94.0 & 96.5 & 18.4 & 18.2 & 18.8 \\
    275 & 22.4 & 77.4 & 82.4 & 58.4 & 95.1 & 97.0 & 20.9 & 20.9 & 20.6 \\
    325 & 36.5 & 80.8 & 84.5 & 74.5 & 96.3 & 97.5 & 23.1 & 23.2 & 23.1 \\
    375 & 44.4 & 82.8 & 86.4 & 76.3 & 97.2 & 98.0 & 26.6 & 26.6 & 26.0 \\
    425 & 53.9 & 84.7 & 88.2 & 83.1 & 97.8 & 98.3 & 30.7 & 29.2 & 30.5 \\
    475 & 76.7 & 89.3 & 91.7 & 96.5 & 98.7 & 98.9 & 35.2 & 35.3 & 33.1 \\
    \multicolumn{10}{c}{SVD} \\ 
    25  & 28.5 & 80.4 & 87.2 & 54.2 & 95.9 & 97.8 & 20.6 & 20.8 & 20.7 \\
    75  & 74.4 & 93.4 & 94.4 & 94.8 & 99.0 & 99.3 & 23.9 & 24.3 & 24.2 \\
    125 & 94.5 & 95.6 & 95.9 & 99.5 & 99.6 & 99.5 & 26.6 & 25.7 & 25.6 \\
    175 & 95.6 & 96.0 & 96.2 & 99.6 & 99.6 & 99.6 & 33.8 & 32.9 & 32.7 \\
    225 & 96.0 & 96.1 & 96.2 & 99.6 & 99.6 & 99.7 & 35.3 & 36.4 & 36.9 \\
    275 & 96.1 & 96.2 & 96.2 & 99.6 & 99.7 & 99.7 & 38.6 & 39.1 & 39.2 \\
    325 & 96.1 & 96.3 & 96.3 & 99.6 & 99.7 & 99.6 & 40.3 & 42.5 & 42.3 \\
    375 & 96.2 & 96.3 & 96.3 & 99.6 & 99.7 & 99.6 & 46.3 & 46.2 & 46.2 \\
    425 & 96.2 & 96.3 & 96.3 & 99.6 & 99.7 & 99.6 & 48.1 & 48.5 & 47.8 \\
    475 & 96.2 & 96.3 & 96.3 & 99.6 & 99.7 & 99.6 & 50.5 & 50.3 & 50.4 \\
    \end{tabular}
    \begin{tablenotes}
        \small
        \item Each column of Top-1, Top-3 and Runtime shows the results for the
            models directly after reduction, after three fine-tuning epochs and
            after retraining (30 epochs), in that order, separated by vertical lines.
    \end{tablenotes}
    \end{threeparttable}
\end{table}

\end{document}